*Article*

# Timepoint-Specific Benchmarking of Deep Learning Models for Glioblastoma Follow-Up MRI

**Wenhao Guo and Golrokh Mirzaei ***

Department of Computer Science and Engineering, The Ohio State University, Columbus, OH 43210, USA; guo.2484@buckeyemail.osu.edu
* Correspondence: mirzaei.4@osu.edu

**Simple Summary**

Glioblastoma is an aggressive brain cancer, and follow-up MRI scans are used to determine whether changes after treatment represent real tumor growth or temporary treatment effects. This decision is difficult, especially at the first follow-up. We analyzed 180 patients and compared eleven deep learning models across two follow-up timepoints. Overall accuracy was similar at both timepoints, ranging from about 70% to 74%. However, the second follow-up provided clearer separation between the three clinical outcomes, with the best model improving its F1 score from 0.44 at the first follow-up timepoint to 0.53 at the second follow-up timepoint. A model that combines convolutional features with a state-space sequence method consistently gave the best balance of accuracy and efficiency, while some transformer models reached higher AUC values but required much more computation. These findings offer a practical benchmark to guide future research and clinical tool development.

**Abstract**

Background: Differentiating true tumor progression (TP) from treatment-related pseudoprogression (PsP) in glioblastoma remains challenging, especially at early follow-up. Methods: We present the first timepoint-specific, cross-sectional benchmarking of deep learning models for follow-up MRI using the Burdenko GBM Progression cohort ($n$ = 180). We analyze different post-RT scans independently to test whether architecture performance depends on timepoint. Eleven representative DL families (CNNs, LSTMs, hybrids, transformers, and selective state-space models) were trained under a unified, QC-driven pipeline with patient-level cross-validation. Across both timepoints, accuracies were comparable (~0.70–0.74), but discrimination improved at the second follow-up, with F1 and AUC increasing for several models, indicating richer separability later in the care pathway. Results: A Mamba+CNN hybrid consistently offered the best accuracy–efficiency trade-off, while transformer variants delivered competitive AUCs at substantially higher computational cost, and lightweight CNNs were efficient but less reliable. Performance also showed sensitivity to batch size, underscoring the need for standardized training protocols. Notably, absolute discrimination remained modest overall, reflecting the intrinsic difficulty of TP vs. PsP and the dataset's size and imbalance. Conclusions: These results establish a timepoint-aware benchmark and motivate future work incorporating longitudinal modeling, multi-sequence MRI, and larger multi-center cohorts.

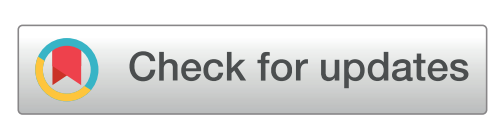

## 1. Introduction

Gliomas are a heterogeneous group of tumors that originate from glial cells in the central nervous system and are among the most common types of brain tumors. They account for approximately 30% of all brain tumors and about 80% of malignant brain tumors. Gliomas can be classified into different types, including astrocytomas, oligodendrogliomas, and glioblastomas, with glioblastoma being the most aggressive and lethal form. The prognosis for patients with glioblastoma remains poor, with a median survival of approximately 15 months despite advances in treatment. The complexity and variability of gliomas, coupled with their infiltrative nature, present significant challenges in diagnosis and treatment, underscoring the need for innovative approaches to improve patient outcomes. A critical challenge in the treatment of brain tumors is differentiating between true tumor progression (TP) and treatment effect (TE), such as pseudoprogression (PsP) and radiation necrosis. TP refers to the actual growth or recurrence of the tumor, indicating a need for a change in treatment strategy to more aggressive or alternative therapies. In contrast, PsP and radiation necrosis are phenomena related to treatment effects and often managed by close monitoring rather than aggressive treatment. Because their imaging appearances can closely mimic true progression, accurate differentiation is difficult and can complicate clinical decision-making.

Machine learning (ML) and deep learning (DL) methods have been increasingly applied to address this challenge. Hu et al. applied a support vector machine on multiparametric MRI, comprising diffusion and perfusion parameters [1]. While the study was limited by a relatively small cohort, their voxel-wise classification approach demonstrated that the integration of advanced MRI modalities enhances discriminative ability. Radiomics-based approaches have also shown promise: a random forest classifier trained on T1C images outperformed neuroradiologists [2], gradient boosting on pre-treatment T1C radiomics demonstrated strong accuracy in external validation [3], and even low-parameter supervised methods yielded moderate AUCs on limited datasets [4]. Building on these early efforts, deep learning frameworks have achieved significant progress. Early studies combined MRI with clinical data in CNN–LSTM frameworks, achieving improved predictive accuracy over unimodal methods [5]. Li et al. proposed a DCGAN–AlexNet framework to mitigate overfitting in small datasets [6], while CNN–LSTM models treating multiparametric MRI as spatial sequences further boosted accuracy and AUC [7]. Automated DL methods, such as 3D DenseNet-121 applied to T2 and CET1 MRI [8], and radiomics classifiers validated on biopsy-proven gliomas [9], confirmed the feasibility of noninvasive PsP–TP prediction. Recent work has moved toward advanced multimodal and transformer-based approaches. Integration of imaging features with MGMT promoter methylation achieved high diagnostic performance [10,11]. Multiparametric MRI combined with MGMT status enhanced PsP–TP classification [12]. CFINet, an attention-based cross-modality feature interaction network leveraging T1- and T2-weighted MRI, demonstrated strong generalization on independent cohorts by effectively capturing complementary information across modalities [13]. Self-supervised multimodal ViTs leveraging contrastive and context-restoration pretraining further improved performance [14]. Nonetheless, most methods remain constrained to single timepoints and heterogeneous datasets, limiting reproducibility and generalizability.

One of the main challenges in ML/DL in this setting is the small sample size and resulting class imbalance due to the low prevalence of TP and PsP. Several studies have used oversampling techniques, such as the Synthetic Minority Oversampling Technique (SMOTE) [15]. Two-network architectures with similarity loss improved robustness under limited-data conditions [16] by training two identical networks in parallel with independently undersampled k-space inputs at different reduction factors. A cosine similarity-based loss combined with L1 loss enforced the networks to learn both common and distinct features, while exponential moving average (EMA) updating of the target network further stabilized training. Methodological cautions highlight that oversampling must be confined to training sets to avoid data leakage [17]. More advanced methods, such as OCH-SMOTE [18], DR-SMOTE [19], and hybrid BSGAN [20], further expanded applicability. Taken together, these findings underscore the importance of class balancing strategies in ensuring robust and generalizable ML/DL pipelines for PsP–TP classification.

Another challenge is how different DL architectures perform across longitudinal follow-ups. Multiple studies have shown that pseudoprogression tends to occur early after radiotherapy; for example, up to 50% of malignant glioma patients show evidence of pseudoprogression on MRI immediately after chemo-radiotherapy [21], and about 60% of PsP cases are reported within the first 3 months post-radiotherapy [22]. By contrast, true tumor progression generally emerges later, and MRI changes beyond 3–6 months are more likely to reflect actual recurrence than transient treatment effects [23]. No prior study has systematically examined whether specific architectures are better suited for early versus late follow-ups—a question with direct clinical relevance.

In this work, we address this critical gap using the Burdenko Glioblastoma Progression Dataset [24], which comprises 180 patients, is substantially larger than previously reported cohorts, and provides longitudinal T1C MRI follow-ups. Table 1 summarizes prior studies, where most datasets ranged between 30 and 130 patients. We preprocessed raw MRI data, mitigated class imbalance through an autoencoder (AE) hybrid with SMOTE and augmentation strategies, and conducted a systematic comparison of deep learning techniques to determine which follow-up timepoint offers greater diagnostic utility and which architectures perform most effectively at each timepoint. We benchmarked CNNs, LSTMs, CNN–LSTM hybrids, ResNets, attention-augmented CNNs, Vision Transformers (2D/3D), Swin Transformers, and Mamba-based hybrids across two clinically relevant timepoints: (1) the first follow-up after radiation therapy and before adjuvant chemotherapy and (2) the second follow-up after combined chemo-radiotherapy. By harmonizing preprocessing protocols and employing patient-level cross-validation, we present the first timepoint-specific comparative evaluation of modern deep learning models in longitudinal glioma imaging. Our findings establish a rigorous benchmark that not only guides methodological choices for machine learning researchers but also delivers translational insights for neuro-oncology, underscoring the importance of timepoint-adaptive modeling strategies in clinical decision support.

**Table 1.** Summary of prior imaging-based machine and deep learning studies for differentiating true progression (TP) from pseudoprogression (PsP) in glioblastoma.

| Reference | # of Patients | # of TP and PsP | Imaging Type | Method | Performance |
|---|---|---|---|---|---|
| Hu et al. [1] | 31 | TP = 15/PsP = 16 | T1, T2, FLAIR, TTP, DWI, DSC, MTT | One-class SVM | AUROC = 0.9439 |
| Qian et al. [25] | 35 | TP = 22/PsP = 13 | DTI | Spatiotemporal discriminative dictionary learning | AUC = 0.875 Acc = 77% |
| Zhang et al. [26] | 79 | TP = 56/PsP = 23 | DTI | SVM | AUC = 0.87 |
| Booth et al. [27] | 26 | TP = 15/PsP = 9 | T2 | SVM | Acc = 86% |
| Ismall et al. [28] | 105 | TP = 34/PsP = 71 | T1, T1C | SVM | AUC = 90.2% |
| Jang et al. [5] | 78 | TP = 48/PsP = 30 | T1, T1C | CNN+LSTM | AUC = 0.83 |
| Li et al. [6] | 84 | TP = 61/PsP = 23 | DTI | DC-AL GAN+SVM | Acc = 92% |
| Kim et al. [29] | 95 | TP = 49/PsP = 46 | T1, T1C, FLAIR, DWI, DSC | Logistic Regression | AUC = 0.96 Acc = 95% |

**Table 1.** *Cont.*

| Reference | # of Patients | # of TP and PsP | Imaging Type | Method | Performance |
|---|---|---|---|---|---|
| Elshafeey et al. [30] | 105 | TP = 83/PsP = 22 | DSE, DSC | SVM | AUC = 0.89 Acc = 90.82% |
| Bani-sadr et al. [31] | 76 | TP = 53/PsP = 23 | T1, T1C, FLAIR | Random forest | AUC = 0.85 Acc = 79.2% |
| Lee et al. [7] | 23 | --- | T1, T1C, T2, T2FLAIR, DWI, T1-post-T1-minus pre-contrast, T2 minus FLAIR | CNN+LSTM | AUC = 0.81 |
| Jang et al. [32] | 104 | TP = 66/PsP = 38 | T1C | CNN | AUC = 0.86 |
| Liu et al. [33] | 84 | TP = 61/PsP = 23 | DTI | CNN | AUC = 0.98 Acc = 88% |
| Akbari et al. [34] | 83 | TP = 63/PsP = 20 | T1, T1C, T2, T2FLAIR, DTI, DSC | SVM | AUC = 0.919 Acc = 87.3% |
| Lohmann et al. [35] | 34 | TP = 18/PsP = 16 | FET-PET | Random forest | AUC = 0.73 Acc = 70% |
| Kebir et al. [36] | 44 | TP = 30/PsP = 14 | FET-PET | Linear discriminant analysis | AUC = 0.93 |
| Sun et al. [2] | 77 | TP = 51/PsP = 26 | T1C | Random forest | Acc = 72.78% |
| Barine et al. [37] | 35 | PsP = 8/ Other = 27 | T1C | Random forest | AUC = 0.82 |
| Moassefi et al. [8] | 124 | TP = 61/PsP = 63 | T1, T2 | 3D-DenseNet | AUC = 0.756 Acc = 76.4% |
| Ari et al. [3] | 131 | TP = 64/PsP = 67 | T1C | Generalized boosted regression models | AUC = 0.915 Acc = 76.04% |
| McKenney et al. [11] | 74 | TP = 57/PsP = 17 | T1 | Recursive feature elimination random forest classifier | AUC = 0.6 |
| Warner et al. [4] | 50 | TP = 37/ PsP = 13 | T1, T1C, T2, T2FLAIR | Geographically weighted regression | AUC ≈ 0.6 |
| Yadav et al. [3] | 75 | TP = 55/ PsP = 20 | T1, T1C, T2FLAIR | SVM | Acc = 85% |
| Lv et al. [13] | 52 | relapse = 42/ PsP = 10 | T1, T2 | CFINet | AUC = 0.929 Acc = 95.4% |
| Gomaa et al. [14] | 79 | TP = 45/ PsP = 34 | T1C, T2FLAIR | ViT | AUC = 0.753 Acc = 75% |
| Wang et al. [38] | 114 | TP= 69/ PsP= 45 | T1C | 3D-CNN | AUC = 0.74 |

# means number, DTI = diffusion tensor imaging, T1 = T1-weighted imaging, T1C = T1-weighted contrast-enhanced imaging, T2 = T2-weighted imaging, FLAIR = fluid-attenuated inversion recovery, DTI = diffusion tensor imaging, DSC = dynamic susceptibility contrast perfusion MRI, DSE = diffusion-sensitive echo imaging, TTP = time-to-peak, MTT= mean transit time, FET-PET = fluoroethyl-tyrosine PET, Acc = accuracy.

## 2. Methods

We systematically developed and benchmarked a diverse set of architectures to interrogate post-treatment glioblastoma MRI, spanning conventional 3D CNNs and ResNets, sequential models (LSTM-based), transformer variants (2D/3D-ViT, Swin Transformer), Mamba models, and novel hybrids that combine CNNs with attention, LSTM, shift windows patching, or Mamba state-space modules. The detailed architectures of these models are provided in the Supplementary Materials.

To systematically investigate how model architecture and clinical timepoint affect predictive performance, we developed a standardized pipeline encompassing dataset curation, preprocessing, model implementation, training/validation, and evaluation. The workflow was designed to ensure reproducibility, prevent information leakage, and provide a fair basis for comparison across architectures.

*2.1. Dataset and Preprocessing*

We used the Burdenko Glioblastoma Progression Dataset (Burdenko-GBM-Progression), https://www.cancerimagingarchive.net/collection/burdenko-gbm-progression/ (accessed on 2 December 2025. Archived version: https://web.archive.org/web/20240101000000/https://www.cancerimagingarchive.net/collection/burdenko-gbm-progression/), comprising 180 patients with primary glioblastoma treated between 2014 and 2020. The dataset includes multi-sequence MRI, CT, clinical, and molecular data; for this study, we focused exclusively on contrast-enhanced T1-weighted MRI (T1C), the most widely available and clinically relevant sequence for follow-up assessment. Imaging was acquired on scanners from four vendors with heterogeneous protocols, reflecting real-world clinical variability. Notably, the dataset documentation does not report scanner field strengths. The dataset is publicly available through The Cancer Imaging Archive (TCIA).

All T1C volumes were standardized through a multi-timepoint pipeline (Figure 1). First, raw DICOM series were converted to NIfTI format using dicom2nifti 2.6.1. and subjected to quality control. Implausible acquisitions were excluded using geometry thresholds on voxel spacing, anisotropy, and slice thickness; among valid candidates, the volume with the highest clarity score was selected per timepoint. Logs and slice previews were generated for auditability. Retained volumes were resampled to $128^3$ voxels, skull-stripped using the BET2 (FSL 6.0.7.17, Oxford, UK), rigidly registered to MNI152 space with the "MNI152_T1_1mm.nii.gz" template provided in FSL 6.0.7.17, and Z-score normalized within the brain mask to ensure cross-subject consistency. Deterministic image–label pairing ensured reproducibility at the patient level. For training, data imbalance was mitigated through oversampling and augmentation. Synthetic 3D MR volumes were generated using a lightweight autoencoder with latent-space interpolation, and mild perturbations—small affine transforms (in-plane rotations within $\pm 3°$ and scaling between 0.98 and 1.02) and Gaussian noise (zero-mean, $\sigma \leq 0.02$)—were applied to both real and synthetic data. After augmentation, 1611 subjects were available at the first follow-up and 1431 subjects at the second follow-up. Full implementation details, including thresholds and parameter settings, are provided in the Supplementary Methods.

*2.2. Labeling Strategy*

In the Burdenko dataset, each follow-up visit is annotated with a clinical label, which can vary over time. To create a single, reproducible outcome label per patient, we developed a consolidation framework in consultation with a neuro-oncologist. The rules were as follows:

- Progression override: Any evidence of true progression at any timepoint defined the patient as Progression, irrespective of prior or subsequent labels, given its clinical impact.
- Pseudoprogression: A patient was labeled PsP if the most recent follow-up indicated PsP and no prior imaging confirmed progression. PsP was also retained if the initial PsP was followed by stability or response without subsequent progression.
- Stable disease: Patients with $\geq$3 consecutive follow-ups showing only stability or response, without new progression, were labeled Stable.
- Scarce follow-up: For patients with $\leq$2 assessments, the most recent report determined the label (PsP if the last scan was PsP; otherwise, Progression).
- Distant progression: Cases with new lesions outside the primary site were immediately classified as Progression.

This systematic framework ensured that final labels were clinically meaningful, consistent across patients, and reproducible for downstream modeling. Accordingly, the final consolidated outcome categories consisted of three labels: Progression, Pseudoprogression, and Stable.

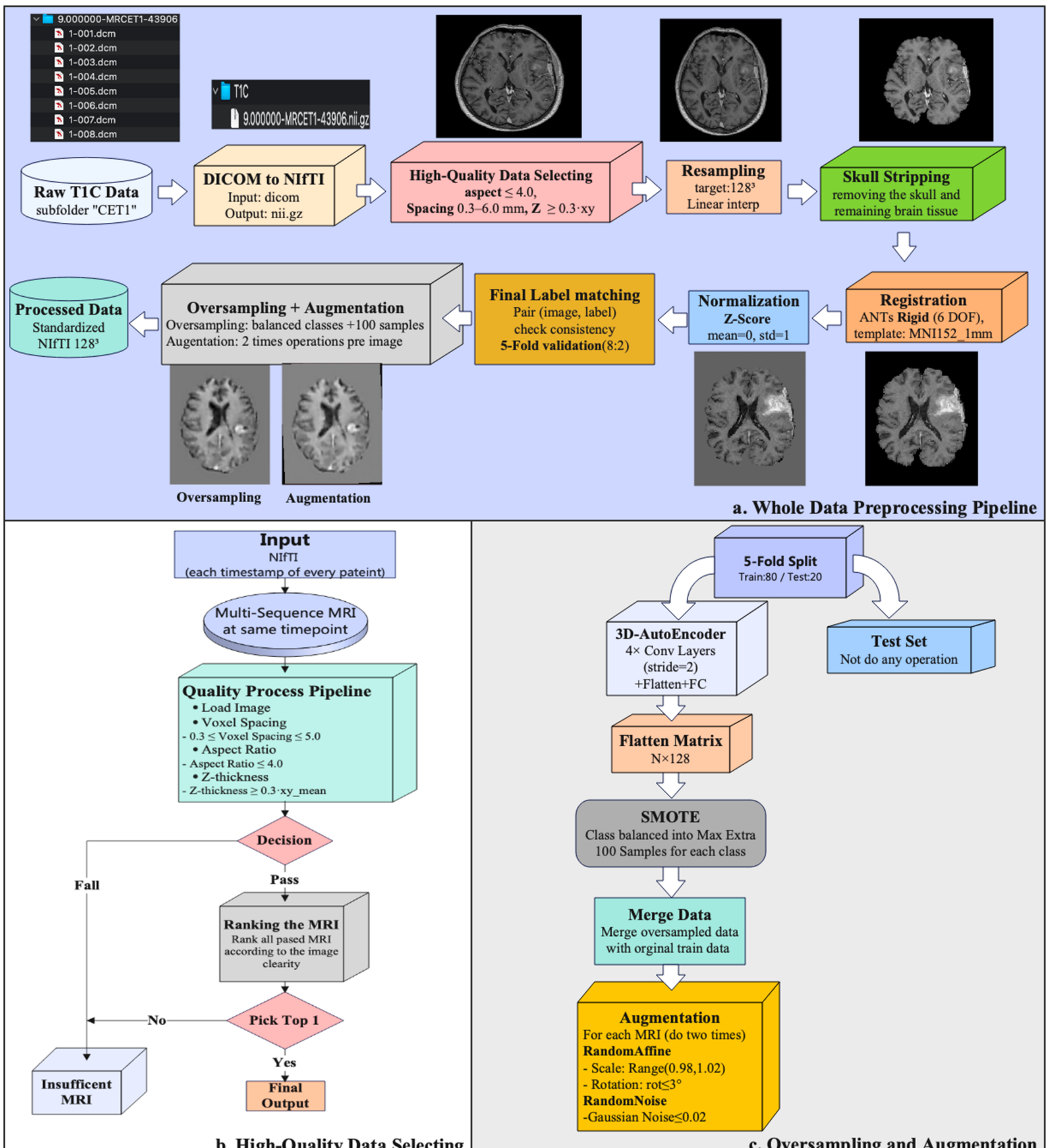

**Figure 1.** Preprocessing workflow. (**a**) Whole-set pipeline from raw T1C identification to standardized outputs. (**b**) High-quality selection logic per patient–timepoint with explicit geometry thresholds and clarity-score ranking; per-series CSV logging and preview generation. (**c**) Training-only oversampling and augmentation: 3D AE latent-space SMOTE with decoding to 3D volumes augmentations; validation/test splits remain unmodified.

## 2.3. Model Architectures

We benchmarked eleven representative DL architectures spanning three design families: (1) Base volumetric models: 3D CNN, LSTM, 3D-ViT, and ResNet. These capture spatial or sequential dependencies directly from T1C volumes. (2) Hybrid models: CNN+LSTM, CNN+SE Attention, and 2D-ViT+LSTM. These combine local convolutional features with sequential or attention-based mechanisms. (3) Advanced models: 3D Swin Transformer, Swin CNN, 2D-Mamba, and 2D-Mamba+CNN, which incorporate hierarchical attention or state-space modeling for efficient long-range dependency capture. Architectures were chosen to represent the spectrum of contemporary strategies: convolutional inductive biases, recurrent modeling of slice order, attention-based global context, and state-space efficiency. All models were trained under the same preprocessing, augmentation, and patient-level cross-validation pipeline, ensuring that observed performance differences reflected architecture rather than implementation. The architectures of the base volumetric models and hybrid models are provided in Supplementary Figures S1–S7 Methods. The details of the developed models for the advanced models are described as follows:

**Shift Windows Transformer (Swin Transformer)—**To reduce the quadratic cost of global attention, Swin Transformer introduces windowed self-attention with shifted windows and a hierarchical (pyramidal) design, yielding strong accuracy–efficiency trade-offs. For volumetric MRI, the input T1C volume ($1 \times 128 \times 128 \times 128$) was partitioned into non-overlapping $4 \times 4 \times 4$ patches and linearly embedded, as shown in Figure 2a. The token grid was processed by four 3D Swin stages (W-MSA/SW-MSA+MLP with residual connections and layer normalization). Channel widths increased $48 \rightarrow 96 \rightarrow 192 \rightarrow 384$ with patch-merging down-sampling the resolution $32^3 \rightarrow 16^3 \rightarrow 8^3 \rightarrow 4^3$, enabling cross-window interaction while controlling compute. The final feature map was aggregated by AdaptiveAvgPool3D (1, 1, 1), projected to 128 dimensions, and passed to a linear–SoftMax classifier ($128 \rightarrow 3$) to yield class probabilities. This formulation captures long-range volumetric dependencies efficiently while preserving locality through windowed attention.

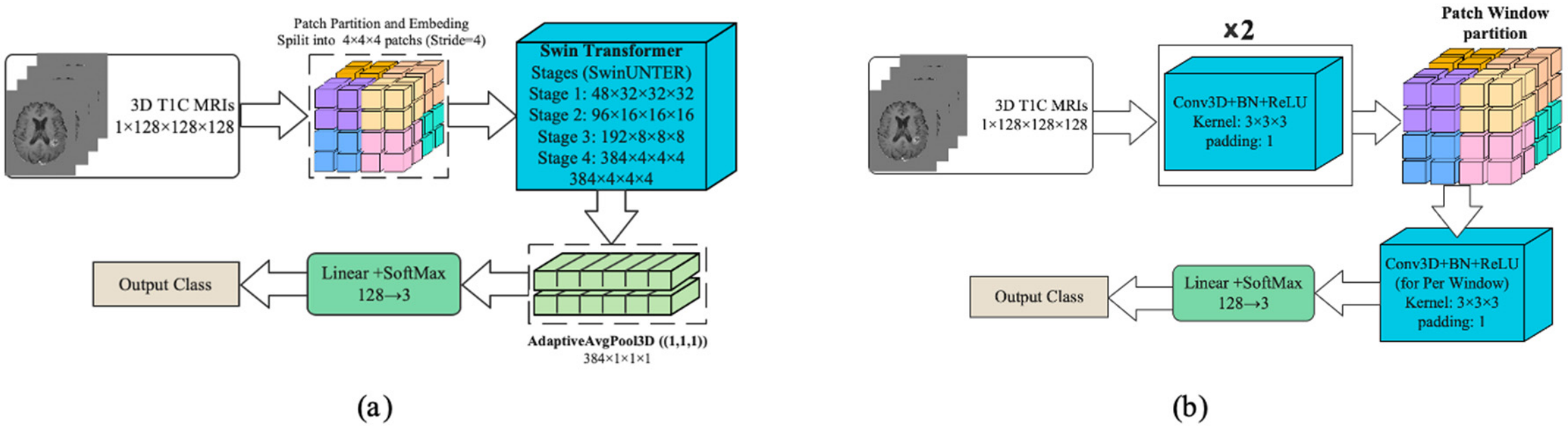

**Figure 2.** The architecture of the developed Swin-based models: (**a**) 3D-Swin Transformer and (**b**) Swin CNN (architectures).

**Shift Window Patching CNN (Swin CNN)—**Despite recent transformer variants, convolutional networks remain highly efficient for volumetric data because shared $3 \times 3 \times 3$ kernels capture local structures with modest parameters. We therefore adopted a 3D-CNN that borrows the window shift idea from Swin: the input T1C volume ($1 \times 128 \times 128 \times 128$) was first processed by two Conv3D–BatchNorm–ReLU blocks (kernel $3^3$, padding 1) to extract low-level features, which is shown in Figure 2b. The resulting feature map was then partitioned into fixed-size 3D windows, within which an additional per-window Conv3D–BN–ReLU was applied. To allow limited cross-window interaction while preserving locality, a second pass with shifted windows (offset by half the window size) could be performed; in practice, this yielded stable performance without materially increasing memory. Window-local convolutions focused capacity on fine anatomical details and constrained computation/memory, while the initial global blocks provided broader context. Features were aggregated (global pooling to a 128-D vector) and fed to a linear–SoftMax head ($128 \rightarrow 3$) to produce class probabilities.

**Two-Dimensional Linear-Time Sequence Modeling with Selective State Spaces (2D-Mamba)—**Mamba is a selective state-space model (SSM) introduced for long-sequence modeling with linear-time complexity; vision adaptations (often called Vision Mamba) map 2D grids to scan orders processed by SSM blocks, offering competitive accuracy with fewer parameters than attention-based transformers. In Figure 3a, each preprocessed T1-contrast (T1C) volume was represented as an ordered set of 50 axial slices. Slices were pad/crop–normalized to 128 × 128 and then resized to 224 × 224 and channel-expanded (grayscale replicated) to match ImageNet-style pretraining. A Vision-Mamba tiny backbone produced a 384-dimensional embedding per slice, yielding a sequence of shapes 50 × 384. Sequence features were mean-pooled to a volume-level descriptor, projected to 128 dimensions, and passed to a linear–SoftMax classifier to output three class probabilities. By using SSM blocks to capture long-range dependencies within slices while aggregating across slices with lightweight pooling, this model retains global contextual modeling with favorable computational efficiency.

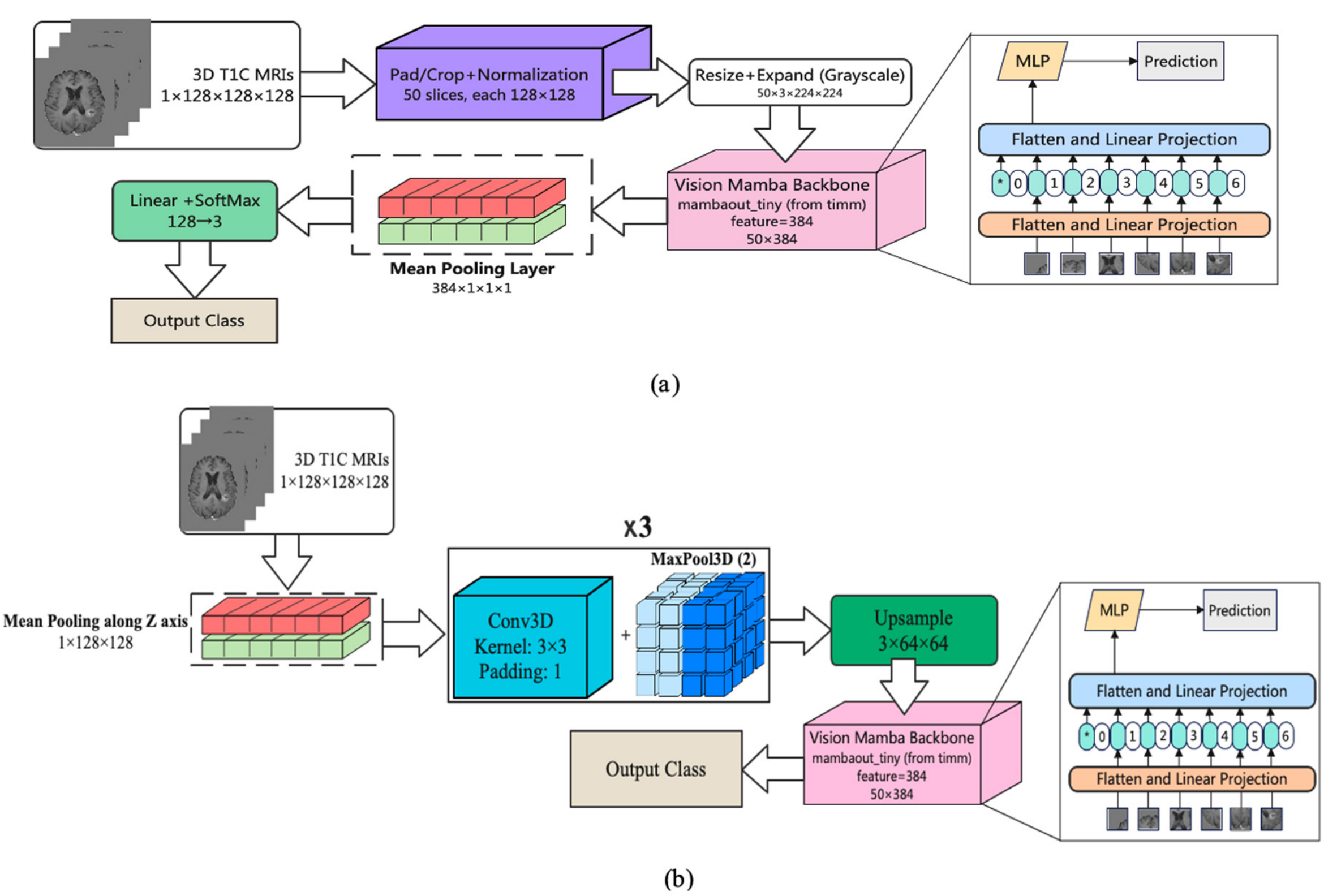

**Figure 3.** The architecture of the developed Mamba-based models: (**a**) 2D Mamba and (**b**) 2D Mamba+CNN architectures. The asterisk (*) denotes the beginning token of the entire sequence.

**Two-Dimensional Mamba Hybrid with CNN—**Convolutional networks capture local spatial patterns efficiently, while Mamba—a selective state-space model (SSM)—models' long-range dependencies in linear time, offering a lighter alternative to self-attention. In Figure 3b, we combined both: each preprocessed T1-contrast volume (1 × 128 × 128 × 128) was first subjected to mean pooling along the z-axis to reduce through-plane redundancy, then passed through three Conv3D–BatchNorm–ReLU blocks interleaved with MaxPool3D (stride 2) to extract and compress volumetric features. The resulting feature map was upsampled and rearranged into an ordered slice sequence (50 slices), which was fed to a Vision-Mamba tiny backbone, producing a 384-dimensional embedding per slice. Sequence embeddings were mean-pooled to a volume-level descriptor, projected to 128 dimensions, and classified with a linear–SoftMax head (128 → 3). This design leverages CNNs for fine local structure while using SSM blocks to aggregate global context across slices with favorable compute and parameter efficiency.

## 3. Results

Analyses were anchored to two clinically defined follow-ups, evaluated as independent cohorts under identical procedures: (1) the first follow-up after radiotherapy (typically ~3–4 weeks post-RT) and (2) the second follow-up after radiotherapy (typically ~2–3 months post-RT). Evaluating the baseline and follow-up scans independently allowed us to characterize model performance across different timepoints. Between the two timepoints of follow-ups, some clear patterns emerged. The performance, in general, was similar across architectures, and the first follow-up was not as good in terms of the separability of classes as the second. The complex models tended to have better discrimination at the expense of significantly larger computation time, while the lightweight models were computationally efficient but less robust. The hybrid architectures achieved the best trade-off between accuracy and efficiency. These patterns form the context for the detailed results reported below.

All experiments were implemented in Python3.9.23 using PyTorch 2.5.1 (with CUDA 12.1) and MONAI 1.4.0. and executed on a workstation with an NVIDIA RTX A6000 (48 GB) GPU and 128 GB RAM. Unless otherwise noted, the models were trained using Adam optimizer (learning rate = $1 \times 10^{-4}$), cross-entropy loss, 10 epochs, determined empirically, and random seeds {21, 33, 42}. Batch sizes of one and eight were used for most models, while computationally intensive architectures (Mamba, Swin Transformer) required smaller batch sizes (one and six) due to GPU memory limits on 3D volumes. The 2D-Mamba+CNN model was further evaluated under batch sizes {1, 2, 4, 8} to assess robustness. Patient-level stratified five-fold cross-validation was used throughout. To prevent information leakage, all preprocessing, augmentation, and synthetic oversampling were fit exclusively on the training split within each fold. Class imbalance was mitigated using latent-space SMOTE (256-D embedding, k = 3) applied to autoencoder features and reconstructed into 3D volumes, with caps on synthetic samples to prevent oversampling bias. Mild augmentations (small affine transforms, Gaussian noise) were applied only to the training data. We report accuracy, macro-averaged F1 score, and macro-averaged AUC to reflect overall performance and robustness under class imbalance. In addition, we measured computational efficiency: FLOPs, parameter count, average batch inference time, and total training runtime.

**Classification performance—**Table 2 summarizes the predictive performance across both follow-up cohorts. Accuracy values were relatively stable across timepoint (≈0.70–0.74), but discrimination improved at the second follow-up, reflected in higher F1 and AUC values for several models. The 2D-Mamba+CNN model demonstrated the most consistent trade-off across follow-ups, achieving 74.5% accuracy and F1 = 0.44 at the first follow-up and improving to 74.1% accuracy and F1 = 0.53 at the second follow-up. This suggests enhanced class separability later in the care pathway. The transformer-based models (3D-ViT, 2D-ViT+LSTM, Swin Transformer) yielded competitive accuracy and AUC but exhibited higher variance and unstable F1 scores. The lightweight CNNs remained efficient and stable, though their discrimination performance lagged behind. Notably, the highest AUC at the first follow-up was obtained with the 3D-ViT model (AUC = 0.57), whereas at the second follow-up, the 2D-Mamba+CNN model achieved the strongest AUC (0.66). Full results across all batch sizes are reported in Supplementary Table S1. Representative predictions from 2D-Mamba+CNN are illustrated in Figure 3b.

**Computational Complexity—**We quantify computational efficiency using FLOPs, trainable parameters, average batch inference time, and total training runtime (Table 3). All values are reported as mean ± SD across seeds {21, 33, 42} under the same hardware configuration. The transformer-based models (2D-ViT+LSTM, 3D-ViT, Swin Transformer) were computationally demanding (≥230 GFLOPs) with runtimes exceeding 1000 min, which challenges clinical deployment despite competitive accuracy/AUC. In contrast, the lightweight CNNs were highly efficient (<13 GFLOPs; <0.1M parameters), with stable runtimes, but exhibited lower discrimination. Full results across all batch sizes are reported in Supplementary Table S2.

**Table 2.** Summary of classification performance (accuracy, F1-score, and AUC) on first and second follow-up MRIs using a consistent batch size (Batch = 1). Complete results across all batch sizes are provided in Supplementary Table S1.

| | **First Follow-Up** | | | **Second Follow-Up** | | |
|---|---|---|---|---|---|---|
| **Model** | **Accuracy** | **F1** | **AUC** | **Accuracy** | **F1** | **AUC** |
| 2DViT+LSTM (Batch = 1) | 0.728 ± 0.0 | 0.2808 ± 0.0 | 0.4612 ± 0.0367 | 0.7024 ± 0.0001 | 0.2749 ± 0.0 | 0.4869 ± 0.0063 |
| 3DViT (Batch = 1) | 0.7181 ± 0.0154 | 0.3607 ± 0.0455 | **0.5728 ± 0.0121** | 0.724 ± 0.0244 | 0.4216 ± 0.0412 | 0.5879 ± 0.0541 |
| CNN (Batch = 1) | 0.6693 ± 0.0216 | 0.3329 ± 0.0209 | 0.4655 ± 0.0738 | 0.6324 ± 0.0645 | 0.3857 ± 0.0778 | 0.5537 ± 0.0322 |
| CNN+Attention (SE) (Batch = 1) | 0.7061 ± 0.0213 | 0.3563 ± 0.0305 | 0.5439 ± 0.0473 | 0.6867 ± 0.0147 | 0.3878 ± 0.0251 | 0.5346 ± 0.0453 |
| CNN+LSTM (Batch = 1) | 0.728 ± 0.0 | 0.2808 ± 0.0 | 0.5257 ± 0.0268 | 0.7057 ± 0.0058 | 0.2903 ± 0.0266 | 0.5412 ± 0.0199 |
| Swin CNN (Batch = 1) | 0.6341 ± 0.0864 | 0.3069 ± 0.0738 | 0.5251 ± 0.0376 | 0.6117 ± 0.084 | 0.2921 ± 0.0257 | 0.4754 ± 0.0118 |
| LSTM (Batch = 1) | 0.7305 ± 0.0043 | 0.2989 ± 0.0157 | 0.5008 ± 0.0723 | 0.7089 ± 0.0113 | 0.2649 ± 0.0174 | 0.4726 ± 0.0134 |
| 2D-Mamba (16 slices) (Batch = 1) | 0.7024 ± 0.0 | 0.2749 ± 0.0 | 0.5556 ± 0.086 | 0.6358 ± 0.1154 | 0.2513 ± 0.0409 | 0.5336 ± 0.0893 |
| 2D-Mamba (50 slices) (Batch = 1) | 0.6095 ± 0.1185 | 0.2413 ± 0.0395 | 0.5198 ± 0.0539 | 0.6706 ± 0.055 | 0.2642 ± 0.0185 | 0.4687 ± 0.0126 |
| 2D-Mamba+CNN (Batch = 1) | **0.7451 ± 0.0261** | **0.4427 ± 0.1143** | 0.5529 ± 0.0493 | **0.7410 ± 0.025** | **0.5264 ± 0.0565** | **0.6332 ± 0.0418** |
| ResNet (Batch = 1) | 0.6935 ± 0.0348 | 0.3331 ± 0.0852 | 0.4981 ± 0.0519 | 0.6705 ± 0.0305 | 0.38 ± 0.0459 | 0.5744 ± 0.015 |
| Swin Transformer (Batch = 1) | 0.7009 ± 0.0535 | 0.2843 ± 0.0198 | 0.4806 ± 0.039 | 0.7056 ± 0.011 | 0.3065 ± 0.0033 | 0.4921 ± 0.0192 |

**Bold** value indicates the best-performing values among all experiments.

The 2D-Mamba+CNN model provided the most favorable efficiency–performance balance: a moderate parameter count (~24M) coupled with very low compute (<1 GFLOP) and reasonable runtime, while maintaining the most consistent accuracy–F1 trade-off across both follow-ups. These characteristics suggest a more practical path toward resource-constrained clinical environments than transformer variants, without the performance drops observed for ultra-light CNNs.

To conclude, the empirical evidence demonstrates that the hybrids of convolutional and Linear-Time Sequence Modeling with Selective State Spaces (Mamba+CNN) consistently yield the best trade-off between predictive cost and computational cost. At both follow-up MRIs, Mamba+CNN had the most stable performance with accuracy and a significantly improved F1-score in the second follow-up, indicating that it could capture spatial and temporal dependencies. While the transformer-based models showed strong accuracy and AUC, their computational requirements prohibited their use in real clinical settings. On the other hand, the efficient CNNs were computationally inexpensive, but the quality of classification dropped significantly. Although Mamba+CNN achieved the best performance across most metrics and performed best with smaller batch sizes, it did not obtain the highest AUC. At the first follow-up, the highest AUC was achieved by 3D-ViT with a batch size of one, while at the second follow-up, CNN+Mamba with a batch size of four achieved the highest AUC. The Mamba+CNN model's predicted results for each class are shown in Figure 4.

**Table 3.** Summary of computational complexity and efficiency metrics (FLOPs, number of parameters, batch time, and runtime) across models using a consistent batch size (batch = 1). Detailed results for all batch sizes are provided in Supplementary Table S2.

| | **First Follow-Up** | | | | **Second Follow-Up** | | | |
|---|---|---|---|---|---|---|---|---|
| **Model** | **FLOPs** | **Params (M)** | **Batch Time (s)** | **Run Time (min)** | **FLOPs** | **Params (M)** | **Batch Time (s)** | **Run Time (min)** |
| 2DViT+LSTM (Batch = 1) | 468.6238 | 5.6421 | 0.373 ± 0.096 | 2142.3667 ± 515.5379 | 468.6238 | 5.6361 | 0.3045 ± 0.0641 | 1766.9167 ± 544.7241 |
| 3DViT (Batch = 1) | 277.2708 | 88.1664 | 0.1753 ± 0.107 | 887.0333 ± 300.3485 | 277.2708 | 88.1664 | 0.1922 ± 0.0301 | 1170.7833 ± 343.5418 |
| CNN (Batch = 1) | 12.3889 | 0.0832 | 0.044 ± 0.0107 | 529.7833 ± 80.0649 | 12.3889 | 0.0832 | 0.021 ± 0.0169 | 468.1633 ± 72.6496 |
| CNN+Attention (SE) (Batch = 1) | 44.821 | 2.4304 | 0.0662 ± 0.0332 | 596.93 ± 64.7776 | 44.821 | 2.4304 | 0.0382 ± 0.0141 | 454.77 ± 32.3068 |
| CNN+LSTM (Batch = 1) | 5.6613 | 1.5707 | 0.0221 ± 0.0133 | 549.25 ± 7.9764 | 5.6613 | 1.2707 | 0.0192 ± 0.0125 | 430.42 ± 48.5683 |
| Swin CNN (Batch = 1) | **0.0255** | **0.0701** | **0.0015 ± 0.0007** | 1114.4133 ± 187.3267 | **0.0255** | **0.0701** | **0.0022 ± 0.0009** | 1024.7733 ± 167.7931 |
| LSTM (Batch = 1) | 7.4189 | 20.3811 | 0.0039 ± 0.0013 | 483.2667 ± 47.298 | 7.4189 | 20.3811 | 0.0033 ± 0.0019 | 411.6133 ± 60.9606 |
| 2D-Mamba (16 slices) (Batch = 1) | 142.9558 | 24.5151 | 0.0808 ± 0.009 | 952.81 ± 248.1051 | 142.9558 | 24.5151 | 0.1279 ± 0.0879 | 1285.73 ± 666.0025 |
| 2D-Mamba (50 slices) (Batch = 1) | 446.7358 | 24.5151 | 0.1544 ± 0.1406 | 2567.8533 ± 242.1994 | 446.7358 | 24.3734 | 0.0846 ± 0.0123 | 671.695 ± 30.1581 |
| 2D-Mamba+CNN (Batch = 1) | 0.7776 | 24.2316 | 0.0801 ± 0.0864 | 787.97 ± 473.7224 | 0.7776 | 24.2316 | 0.0317 ± 0.0169 | **395.4667 ± 68.7862** |
| ResNet (Batch = 1) | 28.4955 | 0.6528 | 0.0784 ± 0.1079 | 511.38 ± 44.35 | 28.4955 | 0.7428 | 0.0169 ± 0.0027 | 416.0133 ± 43.1828 |
| ResNet (Batch = 8) | 28.4955 | 0.6528 | 0.1052 ± 0.0244 | **331.75 ± 21.3322** | 28.4955 | 0.7428 | 0.1382 ± 0.0465 | 896.3633 ± 829.5617 |
| Swin Transformer (Batch = 1) | 236.9559 | 7.8644 | 0.7694 ± 0.1436 | 1324.7833 ± 194.3059 | 236.9559 | 7.8644 | 0.9166 ± 0.3571 | 1404.7267 ± 300.9926 |

**Bold** value indicates the best-performing values among all experiments.

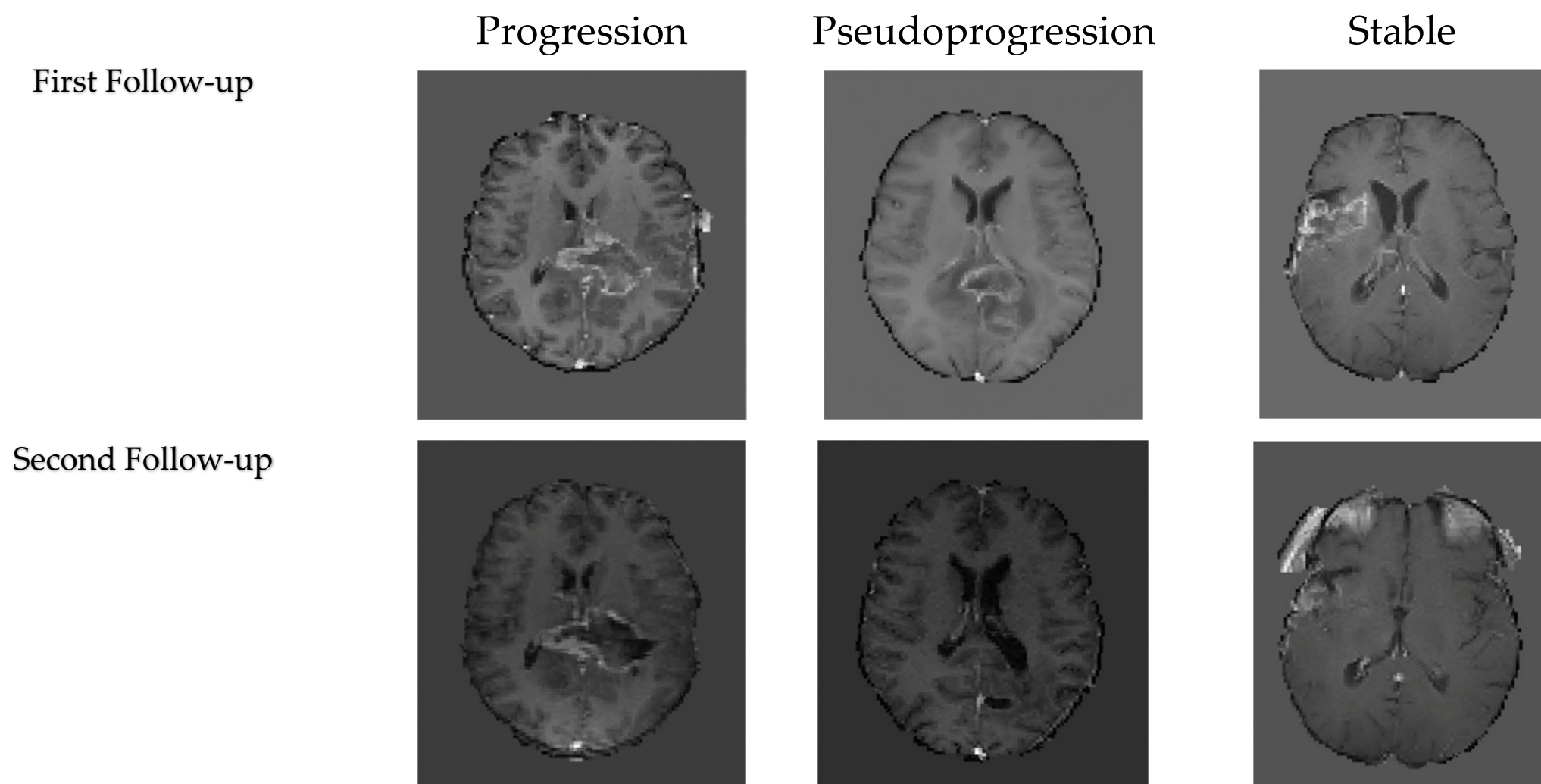

**Figure 4.** Representative examples of model predictions of the first follow-up's and second follow-up's three classes T1C MRI.

## 4. Discussion

In general, the second follow-up exhibited a greater separation between progression, pseudoprogression, and stable disease, representing the expected shift in the clinic from treatment-related changes in early imaging to more definitive evidence of recurrence at later timepoints. Although transformer-based models provide strong discrimination, they are computationally intensive, which restricts their use in daily clinical practice. The lightweight CNNs were highly efficient but offered lower robustness, whereas the 2D-Mamba+CNN model provided the strongest balance between performance and computational cost, making it a more viable candidate for clinical integration.

Several limitations should be noted. This study focused on two standardized post-treatment MRI timepoints that were consistently available across patients. The first follow-up, obtained approximately 3 weeks after radiotherapy (RT), reflects the early post-treatment period when inflammatory and treatment-related effects are most prominent. The second follow-up, performed 2–3 months after RT, represents a later stage in which evolving tumor biology or emerging pseudoprogression becomes more apparent. Although additional MRIs existed for some patients, these timepoints were not uniformly available or aligned across the cohort. Therefore, we selected these two clinically meaningful and consistently acquired follow-ups to ensure comparability and avoid substantial loss of sample size or selection bias. The dataset is still quite small and imbalanced, and multi-center validation is required to verify its generalizability. In addition, this study only assessed T1C MRI. While this sequence is widely accessible, it cannot comprehensively describe treatment-induced alterations on imaging, and the addition of multi-sequence MRI, such as FLAIR, T1, and T2, may lead to an even better separation of progression from pseudoprogression.

## 5. Conclusions

We presented the first timepoint-specific benchmarking of deep learning architectures for differentiating true progression from pseudoprogression in glioblastoma using follow-up MRI. Evaluating eleven model families within a unified, quality-controlled pipeline, we found that overall performance remains moderate but improves at later follow-ups, consistent with richer clinical separability between treatment effects and true recurrence over time. These results establish a timepoint-aware benchmark for follow-up MRI analysis in glioblastoma and underscore the importance of standardized, leakage-aware training and evaluation protocols for this challenging task.

At the architectural level, our findings extend prior work on hybrid models for medical imaging by systematically comparing CNNs, transformers, and state-space-based approaches in longitudinal glioblastoma follow-ups. A 2D-Mamba+CNN model provided the most reliable trade-off between predictive accuracy and computational efficiency, whereas transformer-based models (e.g., ViT, Swin) achieved competitive accuracy and AUC at the cost of higher computational burden and more variable F1 scores. In contrast, lightweight CNNs were highly efficient in terms of FLOPs and parameter counts but exhibited lower discrimination, reflecting the difficulty of handling class imbalance and complex spatiotemporal patterns in this setting. Broader validation on multi-center, multi-sequence cohorts, incorporation of clinical and molecular covariates, and exploration of self-supervised and model-compression strategies will be critical next steps toward practical deployment of automated follow-up MRI analysis in neuro-oncology.

**Supplementary Materials:** The following supporting information can be downloaded at: https://www.mdpi.com/article/10.3390/cancers18010036/s1, Figure S1: CNN architecture; Figure S2: LSTM architecture; Figure S3: 3D-ViT architecture; Figure S4: 3D-ResNet architecture; Figure S5: CNN+LSTM architecture; Figure S6: CNN+SE Attention architecture; Figure S7: 2D ViT+LSTM architecture; Table S1: Summary of classification performance on first and second follow-up MRIs; Table S2: Computational complexity and runtime metrics for all models.

**Author Contributions:** Methodology, formal analysis, original draft preparation, W.G. Supervision, review, and editing, G.M. All authors have read and agreed to the published version of the manuscript.

**Funding:** This research received no funding.

**Institutional Review Board Statement:** Ethical review and approval were waived for this study because the MRI data were fully anonymized and obtained from a publicly available dataset on The Cancer Imaging Archive, and no identifiable human information was accessed.

**Informed Consent Statement:** Patient consent was waived due to the use of fully anonymized and publicly available MRI data from The Cancer Imaging Archive, and no identifiable information was accessed.

**Data Availability Statement:** The MRI data used in this study are publicly available from The Cancer Imaging Archive (TCIA) under the Burdenko Glioblastoma Progression Dataset (Burdenko-GBM-Progression). The dataset can be accessed at https://www.cancerimagingarchive.net/collection/burdenko-gbm-progression/ (accessed on 2 December 2025). Archived version: https://web.archive.org/web/20240101000000/https://www.cancerimagingarchive.net/collection/burdenko-gbm-progression/.

**Acknowledgments:** The authors thank Pierre Giglio, Director of Neuro-Oncology at The Ohio State University Wexner Medical Center, for his advice on MRI labeling and for reviewing the labeling strategy.

**Conflicts of Interest:** The authors declare no conflicts of interest.